\documentclass[pdflatex,sn-mathphys,iicol]{sn-jnl}% Math and Physical Sciences Reference Style
\usepackage{epigraph}
\usepackage{longtable}

\usepackage{textcomp}
\usepackage{setspace}

\usepackage{amssymb}

\usepackage{graphicx}
\usepackage{subcaption}

\usepackage{multirow}

\usepackage{color}
\definecolor{PyOrange}{RGB}{255, 201, 14}
\definecolor{PyBlue}{RGB}{112, 146, 190}

\definecolor{WordGreen}{RGB}{100, 136, 40}
\definecolor{WordDarkGrey}{RGB}{82, 82, 82}
\definecolor{WordRed}{RGB}{192, 80, 77}
\definecolor{WordBlue}{RGB}{0, 122, 192}

\definecolor{WordLightBlue}{RGB}{218, 238, 243}
\definecolor{WordLightGreen}{RGB}{234, 241, 221}

\definecolor{WordFillGreen}{RGB}{194, 214, 155}
\definecolor{WordFillRed}{RGB}{252, 214, 182}
\definecolor{WordFillGray}{RGB}{217, 217, 217}

\UseRawInputEncoding
\usepackage{times}
\usepackage{epsfig}
\usepackage{amssymb}
\usepackage[ruled,vlined,linesnumbered]{algorithm2e}
\usepackage{comment}
\usepackage{array, makecell} 
\usepackage{lettrine}
\usepackage{url} 
\usepackage{lscape}
\usepackage{tabularx}
\usepackage{colortbl}
\usepackage{hhline}
\usepackage{vcell}
\usepackage{longtable}
\usepackage{vcell}
\usepackage{pdflscape}
\usepackage{sidecap}
\usepackage[export]{adjustbox} %
\usepackage{acronym}
\acrodef{FCN}[FCN]{Fully Convolutional Network}
\acrodef{GAME}[GAME]{Grid Average Mean Absolute Error}
\acrodef{DL}[DL]{Deep Learning}
\acrodef{DNN}[DNN]{Deep Neural Network}
\acrodef{ML}[ML]{Machine Learning}
\acrodef{CV}[CV]{Computer Vision}
\acrodef{AI}[AI]{Artificial Intelligence}
\acrodef{CNN}[CNN]{Convolutional Neural Network}
\acrodef{RNN}[RNN]{Recurrent Neural Network}
\acrodef{GAN}[GAN]{Generative Adversarial Network}
\acrodef{JCU}[JCU]{James Cook University}
\acrodef{MAE}[MAE]{Mean Average Error}
\acrodef{MAP}[mAP]{Mean Average Precision}
\acrodef{CA}[CA]{Classification Accuracy}
\acrodef{LCFCN}[LCFCN]{Localization-based Counting loss Fully Convolutional Network}
\acrodef{IoT}[IoT]{Internet of Things}
\acrodef{MLP}[MLP]{Multi-Layer Perceptrons}
\usepackage{xspace}

\newcommand{\wrt}{\emph{w.\thinspace{}r.\thinspace{}t.}\@\xspace}

\DeclareMathOperator*{\argmax}{arg\,max}
\usepackage[capitalize]{cleveref}
\crefname{section}{Sec.}{Section}
\usepackage{arydshln}
\usepackage[switch]{lineno}
\usepackage{listings}
\definecolor{codegreen}{rgb}{0,0.6,0}
\definecolor{codegray}{rgb}{0.5,0.5,0.5}
\definecolor{codepurple}{rgb}{0.58,0,0.82}
\definecolor{backcolour}{rgb}{0.95,0.95,0.92}
\definecolor{cleacolorr}{rgb}{1,1,1}

\lstdefinestyle{mystyle}{
    backgroundcolor=\color{cleacolorr},   
    commentstyle=\color{codegreen},
    keywordstyle=\color{magenta},
    numberstyle=\tiny\color{codegray},
    stringstyle=\color{codepurple},
    basicstyle=\ttfamily\footnotesize,
    breakatwhitespace=false,         
    breaklines=true,                 
    captionpos=b,                    
    keepspaces=true,                 
    numbers=left,                    
    numbersep=5pt,                  
    showspaces=false,                
    showstringspaces=false,
    showtabs=false,                  
    tabsize=2
}

\usepackage{lmodern}     % Use Latin Modern fonts for scalability
\usepackage[T1]{fontenc} % Enable more font shapes
\usepackage{amsmath}     % For advanced math symbols and fonts
\usepackage{mathrsfs}    % For special script fonts (if needed)
\usepackage{type1cm}
\usepackage{anyfontsize}

\begin{document}

\title[Self-Supervised Fish Segmentation]{Overcoming Annotation Bottlenecks in Underwater Fish Segmentation: A Robust Self-Supervised Learning Approach}


\author*[1,2]{\fnm{Alzayat} \sur{Saleh}}\email{alzayat.saleh@my.jcu.edu.au}

\author[1]{\fnm{Marcus} \sur{Sheaves}}%\email{iiauthor@gmail.com}

\author[1,2]{\fnm{Dean} \sur{Jerry}}%\email{iiiauthor@gmail.com}

\author[1,2]{\fnm{Mostafa} \sur{Rahimi~Azghadi}}%\email{iiiauthor@gmail.com}


\affil[1]{College of Science and Engineering, James Cook University, Townsville, QLD, Australia}
\affil[2]{ARC Research Hub for Supercharging Tropical Aquaculture through Genetic Solutions, James Cook University, Townsville, QLD, Australia}

% \affil*[1]{\orgdiv{Department}, \orgname{Organization}, \orgaddress{\street{Street}, \city{City}, \postcode{100190}, \state{State}, \country{Country}}}

% \affil[2]{\orgdiv{Department}, \orgname{Organization}, \orgaddress{\street{Street}, \city{City}, \postcode{10587}, \state{State}, \country{Country}}}

% \affil[3]{\orgdiv{Department}, \orgname{Organization}, \orgaddress{\street{Street}, \city{City}, \postcode{610101}, \state{State}, \country{Country}}}


\abstract{Accurate fish segmentation in underwater videos is challenging due to low visibility, variable lighting, and dynamic backgrounds, making fully-supervised methods that require manual annotation impractical for many applications. This paper introduces a novel self-supervised learning approach for fish segmentation using Deep Learning. Our model, trained without manual annotation, learns robust and generalizable representations by aligning features across augmented views and enforcing spatial-temporal consistency. We demonstrate its effectiveness on three challenging underwater video datasets: DeepFish, Seagrass, and YouTube-VOS, surpassing existing self-supervised methods and achieving segmentation accuracy comparable to fully-supervised methods without the need for costly annotations. Trained on DeepFish, our model exhibits strong generalization, achieving high segmentation accuracy on the unseen Seagrass and YouTube-VOS datasets. Furthermore, our model is computationally efficient due to its parallel processing and efficient anchor sampling technique, making it suitable for real-time applications and potential deployment on edge devices. We present quantitative results using Jaccard Index and Dice coefficient, as well as qualitative comparisons, showcasing the accuracy, robustness, and efficiency of our approach for advancing underwater video analysis}

\keywords{ Computer Vision, Convolutional Neural Networks, Underwater Videos, Deep Learning, Transformer, Self-Supervised Learning.}
% \keywords{ Computer Vision, Convolutional Neural Networks, Image and Video Processing, Underwater Videos, Deep Learning, Transformer, Self-Supervised Learning, Fish Segmentation, Marine Ecology, Aquaculture.}



\maketitle

%%%%%%%%%%%%%%%%%%%%%%%%%%%%%%%%%%%%%%%%%%%%%%%%%%%%%%%%%%%%%%%%

% ############################################
\section{Introduction} \label{secintro}
The health and sustainability of aquatic ecosystems and aquaculture operations depend on effective monitoring and management of fish populations. Fish segmentation, the process of automatically identifying and outlining fish in images or videos, is crucial for estimating fish abundance, tracking movement patterns, analyzing behavior, and assessing fish stock health. However, accurately segmenting fish in underwater environments is challenging due to low visibility, variable lighting, dynamic backgrounds, and the diverse appearances and movements of fish.

Traditional computer vision approaches struggle with these complexities, and fully supervised deep learning methods, while effective, require extensive manually annotated datasets. Creating such datasets is time-consuming, expensive, and prone to biases, limiting scalability and generalizability. For instance, segmenting a single fish in an image can take approximately two minutes, leading to significant annotation costs for even moderately sized datasets. Unsupervised and self-supervised learning methods have emerged as promising alternatives, eliminating the need for costly annotations by leveraging inherent structures and relationships within unlabeled data \cite{saleh2022a}.

\begin{figure}[t]
\centering
\includegraphics[width=0.9\linewidth]{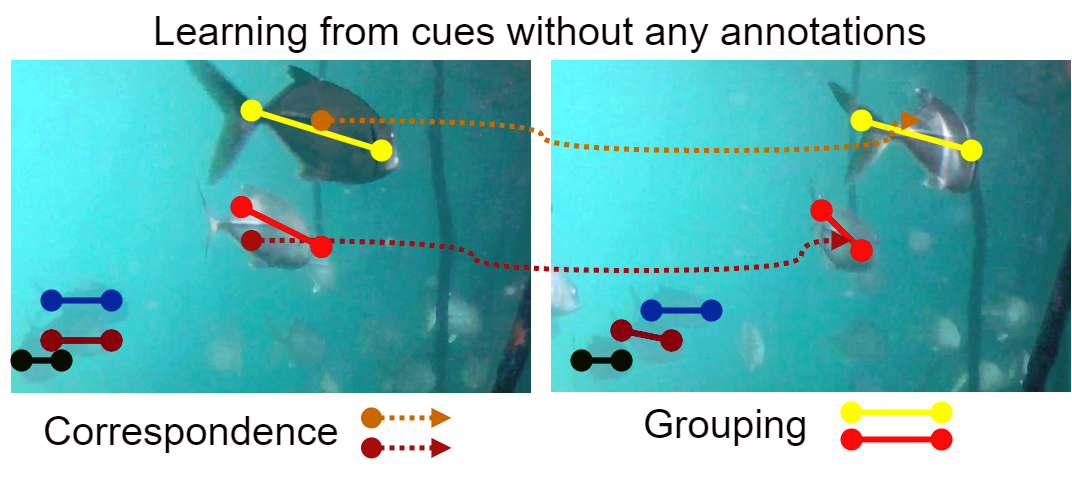}
\caption{The natural visual artefact dynamics provide important cues about the composition of scenes and how they change.}
\label{fig:9}
\end{figure}

Contrastive learning, a prominent self-supervised technique, has shown success in visual representation learning by distinguishing between positive (similar) and negative (dissimilar) pairs of augmented views \cite{Shao2021}. Similarly, correspondence learning leverages temporal coherence and spatial relationships in video data to propagate information and segment objects effectively over time \cite{lai2019corrflow,Jabri2020}. Vision Transformers (ViTs) have also gained traction in computer vision tasks, capturing long-range dependencies and global context through self-attention mechanisms \cite{Dosovitskiy2020}. These advancements provide a strong foundation for addressing the challenges of underwater fish segmentation.

To address these challenges, this paper introduces a novel self-supervised learning approach for robust and efficient fish segmentation in underwater videos. Our method leverages Transformer networks, which excel at capturing long-range dependencies and global context in computer vision tasks. Unlike supervised methods, our approach eliminates the need for manual annotation by learning powerful representations directly from unlabeled videos. This is achieved through feature alignment across augmented views and enforcing spatial-temporal consistency, enabling the model to focus on underlying fish features and maintain accuracy in dynamic scenes, see \cref{fig:9}.

\noindent
Our approach offers several key advantages:
\begin{itemize}
    \item \textbf{Eliminates Annotation Bottleneck}: Removes the need for manual annotation, reducing time and cost for data preparation.
    \item \textbf{Enhanced Generalizability}: Learns robust representations from diverse unlabeled data, improving performance on unseen datasets.
    \item \textbf{Computational Efficiency}: Employs an efficient architecture suitable for real-time applications and resource-constrained devices.
\end{itemize}

We demonstrate the effectiveness of our approach on three challenging datasets: DeepFish \cite{Saleh2020a}, Seagrass \cite{Ditria2021a}, and YouTube-VOS \cite{Xu2018b}. Trained solely on DeepFish, our model generalizes well to Seagrass and YouTube-VOS, outperforming existing self-supervised methods and rivaling fully-supervised approaches without costly annotations.

%%%%%%%%%%%%%%%%%%%%%%%%%%%%%%%%%%%%%%%%%%%%%%%%%%%%%%%%%%%%%%%%
\begin{figure*}[t]
\centering
\includegraphics[width=0.80\textwidth]{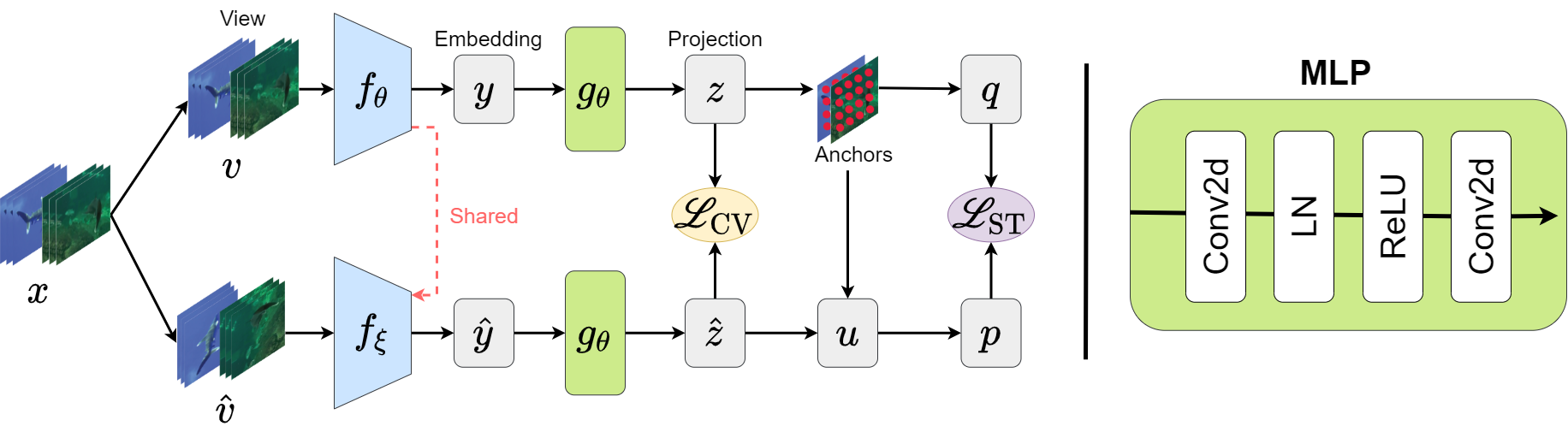}
\caption{
Our proposed framework consists of a single   feature extractor that processes video sequences.
Given a batch of unlabeled video sequences $x$, two batches of different views $v$ and $\hat{v}$ are produced and are then encoded into embeddings $y$ and $\hat{y}$ through the main branch $f_{\theta}$ and the second regularising branch $f_{\xi}$, respectively.
The embeddings are fed to a multilayer perceptron (MLP) $g_{\theta}$ to produce the projections $z$ and $\hat{z}$ to compute the cross-view consistency loss $\mathcal{L}_{\text{CV}}$.
The self-training loss $\mathcal{L}_{\text{ST}}$ learns space-time embeddings between the anchors $q$ and pseudo labels $p$ (arg max of $u$, affinities of $\hat{z}$ \wrt anchors.). 
The two branches are identical in architecture with shared weights. 
% The encoders $f$ are CoaT Transformer \cite{xu2021coat} backbones.
}
\label{fig:6}
\end{figure*}
%%%%%%%%%%%%%%%%%%%%%%%%%%%%%%%%%%%%%%%%%%%%%%%%%%%%%%%%%%%%%%%%

% ############################################

%%%%%%%%%%%%%%%%%%%%%%%%%%%%%%%%%%%%%%%%%%%%%%%%%%%%%%%%%%%%%%%%
\begin{figure}[t]
\centering
\includegraphics[width=0.49\textwidth]{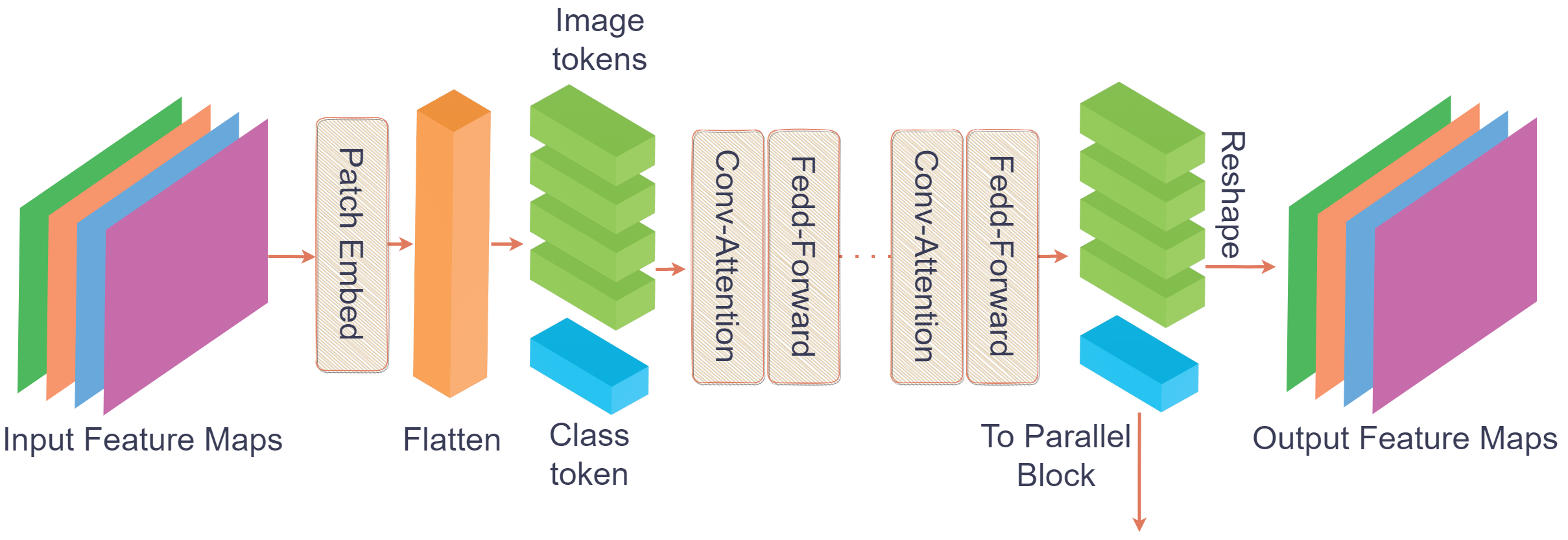}
\caption{
Schematic graph of the serial block in CoaT Transformer  \cite{xu2021coat}. 
Input feature maps are first down-sampled by a patch embedding layer and then flatten the reduced feature maps into a sequence of image tokens.
Multiple Conv-Attention and Feed-Forward layers process the tokenized features, along with a class token (a vector to achieve image classification).
}
\label{fig:7}
\end{figure}
%%%%%%%%%%%%%%%%%%%%%%%%%%%%%%%%%%%%%%%%%%%%%%%%%%%%%%%%%%%%%%%%

\section{Method} \label{secmethod}
This section outlines the architecture and training procedure of our proposed model, which comprises two branches: a main branch and a regularizing branch. Both branches share the same architecture and weights to ensure consistent feature representation learning. The model processes unlabeled video sequences as input, as illustrated in Figure \ref{fig:6}.

\subsection{Model Architecture}
\begin{itemize}
    \item \textbf{Input}: The model receives a batch of unlabeled video sequences.
    \item \textbf{View Augmentation}: Two distinct views of each input frame batch are generated. The main branch processes the original frames, while the regularizing branch operates on augmented versions (random cropping and flipping) to promote invariance to transformations.
    \item \textbf{Feature Encoder (CoaT Transformer)}: The CoaT Transformer \cite{xu2021coat} is used as the feature encoder backbone. It captures both local and global image features through its co-scale mechanism, enabling multi-scale representations for improved segmentation accuracy.
    \item \textbf{Feature Embedding (MLP)}: The CoaT encoder output is passed through a multilayer perceptron (MLP) to produce compact feature embeddings. The MLP consists of two \texttt{Conv2d} layers with Layer Normalization and ReLU activation, reducing the feature dimensionality from 512 to 128.
\end{itemize}

% \subsection{Co-Scale Conv-Attention} \label{seccoat}
% The CoaT Transformer consists of two submodules:
% \begin{itemize}
%     \item \textbf{Conv-Attentional Image Transformer (CAIT)}: Uses a spatial transformer network and convolutional operations to produce a co-scale feature pyramid from a single input image. It also employs convolutions for relative position embeddings in the factorized attention mechanism.
%     \item \textbf{Co-Scale Feature Attention Network (CFAN)}: Dynamically selects informative image parts, enabling the model to encode relevant areas and ignore others. CFAN includes serial and parallel blocks that introduce fine-to-coarse, coarse-to-fine, and cross-scale information into image transformers.
% \end{itemize}
% Given an input image \( I \in \mathbb{R}^{H \times W \times C} \), the serial block downsamples the image features into four resolutions:
% $
% F_1 \in \mathbb{R}^{\frac{H}{4} \times \frac{W}{4} \times C_{1}}, 
% \quad
% F_2 \in \mathbb{R}^{\frac{H}{8} \times \frac{W}{8} \times C_{2}},
% \quad
% F_3 \in \mathbb{R}^{\frac{H}{16} \times \frac{W}{16} \times C_{3}}, 
% \quad
% F_4 \in \mathbb{R}^{\frac{H}{32} \times \frac{W}{32} \times C_{4}}.
% $
% CFAN produces multi-scale feature attention maps efficiently, making it computationally scalable and flexible compared to existing multi-resolution frameworks.

% \color{blue}
\subsection{Co-Scale Conv-Attention} \label{seccoat}
The CoaT Transformer consists of two key submodules: the \textbf{Conv-Attentional Image Transformer (CAIT)} and the \textbf{Co-Scale Feature Attention Network (CFAN)}. These submodules work together to capture both local and global features, enabling the model to handle the complexities of underwater fish segmentation.

\begin{itemize}
    \item \textbf{Conv-Attentional Image Transformer (CAIT)}: 
    The CAIT module combines the strengths of convolutional operations and self-attention mechanisms. It uses a \textbf{spatial transformer network} to process the input image and generate a \textbf{co-scale feature pyramid}. This pyramid captures features at multiple resolutions, allowing the model to focus on both fine-grained details (e.g., fish scales, fins) and broader contextual information (e.g., fish shape, background). The CAIT also employs \textbf{convolutional relative position embeddings} in its factorized attention mechanism, which helps the model understand spatial relationships between different parts of the image. This is particularly useful in underwater scenes, where fish may appear at various scales and orientations.

    \item \textbf{Co-Scale Feature Attention Network (CFAN)}: 
    The CFAN module dynamically selects informative regions of the image, enabling the model to focus on relevant areas while ignoring less important ones. It introduces \textbf{fine-to-coarse}, \textbf{coarse-to-fine}, and \textbf{cross-scale information} into the image transformer through a combination of serial and parallel blocks, see \cref{fig:7}. These blocks process the image at multiple resolutions, ensuring that the model can capture both high-level semantic information and low-level texture details. For example, given an input image \( I \in \mathbb{R}^{H \times W \times C} \), the serial block downsamples the image features into four resolutions:
$
F_1 \in \mathbb{R}^{\frac{H}{4} \times \frac{W}{4} \times C_{1}}, 
\quad
F_2 \in \mathbb{R}^{\frac{H}{8} \times \frac{W}{8} \times C_{2}},
\quad
F_3 \in \mathbb{R}^{\frac{H}{16} \times \frac{W}{16} \times C_{3}}, 
\quad
F_4 \in \mathbb{R}^{\frac{H}{32} \times \frac{W}{32} \times C_{4}}.
$
    This multi-scale approach ensures that the model can effectively segment fish in diverse underwater environments, even when they are partially occluded or appear in dynamic backgrounds.
\end{itemize}

By combining these submodules, the CoaT Transformer captures \textbf{long-range dependencies} and \textbf{global context} through its self-attention mechanisms, while also preserving fine-grained local details through convolutional operations. This makes it particularly well-suited for underwater fish segmentation, where the model must handle variable lighting, low visibility, and complex backgrounds.
% \color{black}

\subsection{Multilayer Perceptron (MLP)} \label{secmlp}
The MLP processes the feature encoder output to generate feature embeddings, reducing the dimensionality from 512 to 128. It consists of two \texttt{Conv2d} layers, with the first layer followed by Layer Normalization and ReLU activation. The PyTorch-like implementation of the MLP is shown in Algorithm \ref{fig_mlp}.

\subsection{Regularizing Branch}
The regularizing branch processes augmented versions of the input frames to avoid degenerate solutions, such as encoding positional cues into feature representations \cite{Jabri2020}. This ensures robust and generalizable feature learning.

\begin{algorithm}[ht]
  \scriptsize
  \caption{Multilayer Perceptron (MLP) Forward Pass} 
  \LinesNumbered
  \KwIn{Input feature tensor $x \in \mathbb{R}^{B \times C \times H \times W}$, where $B$ is the batch size, $C$ is the number of input channels, and $H$, $W$ are the height and width of the feature map.}
  \KwOut{Output feature embeddings $z \in \mathbb{R}^{B \times C' \times H \times W}$, where $C'$ is the number of output channels.}
  
  \For{Each layer in MLP}{
    Apply Conv2d layer with kernel size $1 \times 1$ and stride $1$; \\
    Apply Layer Normalization to the output; \\
    Apply ReLU activation function; \\
    Update feature tensor $x$ with the output of the current layer;
  }
  \Return{Processed feature embeddings $z$}
  \label{fig_mlp}
\end{algorithm}
% \color{black}

\subsection{Anchor Sampling} \label{secgrd}
To improve training efficiency and reduce computational overhead, we define a spatially invariant grid of size \( n \times n \) on the feature tensor from the main branch \( z \), obtaining \( n^2 \) \( z \)-dimensional feature embeddings. One sample per grid cell is selected as an anchor \( k \), ensuring spatial distinctness and full coverage of the feature embeddings. These anchors are shared with the regularizing branch.

Instead of computing pairwise distances between all feature vectors in the batch, we calculate the cosine similarities between the embeddings of the anchors \( k \) and the current features \( z \) using:
\begin{equation}
q_{i,j} = \frac{\exp ( z_i \cdot k_j / \tau ) }{\sum_l \exp (z_i \cdot k_l / \tau) },
\label{eq:affinity}
\end{equation}
where \( \tau \in \mathbb{R}^{+} \) is a temperature hyperparameter, \( z \) represents features from the main branch, \( k \) represents the anchors, \( l \) indexes batch samples, and \( i,j \) index the vector dimension.

For the regularizing branch features, only the predominant anchors are selected. The cosine similarities are calculated as:
\begin{equation}
p_{i} = \argmax_{j \in \mathcal{N}(i)} \frac{\exp ( \hat{z}_i \cdot k_j / \tau ) }{\sum_l \exp (\hat{z}_i \cdot k_l / \tau) },
\label{eq:pseudo_label}
\end{equation}
where \( \mathcal{N}(i) \) is the index set of anchors from the same video clip as the feature vector \( \hat{z}_i \), and \( \hat{z} \) represents features from the regularizing branch.

Equation \ref{eq:affinity} leverages features from multiple videos in the training batch, while Equation \ref{eq:pseudo_label} focuses on features from the same video sequence. This dual approach enables the framework to learn both intra-video and inter-video feature embeddings, preserving fine-grained correspondence associations and instance-level feature discrimination.

 %%%%%%%%%%%%%%%%%%%%%%%%%%%%%%%%%%%%%%%%%%%%%%%%%%%%%%%%%%%%%%%% 
\begin{figure}[t]
\centering
\includegraphics[width=0.95\linewidth]{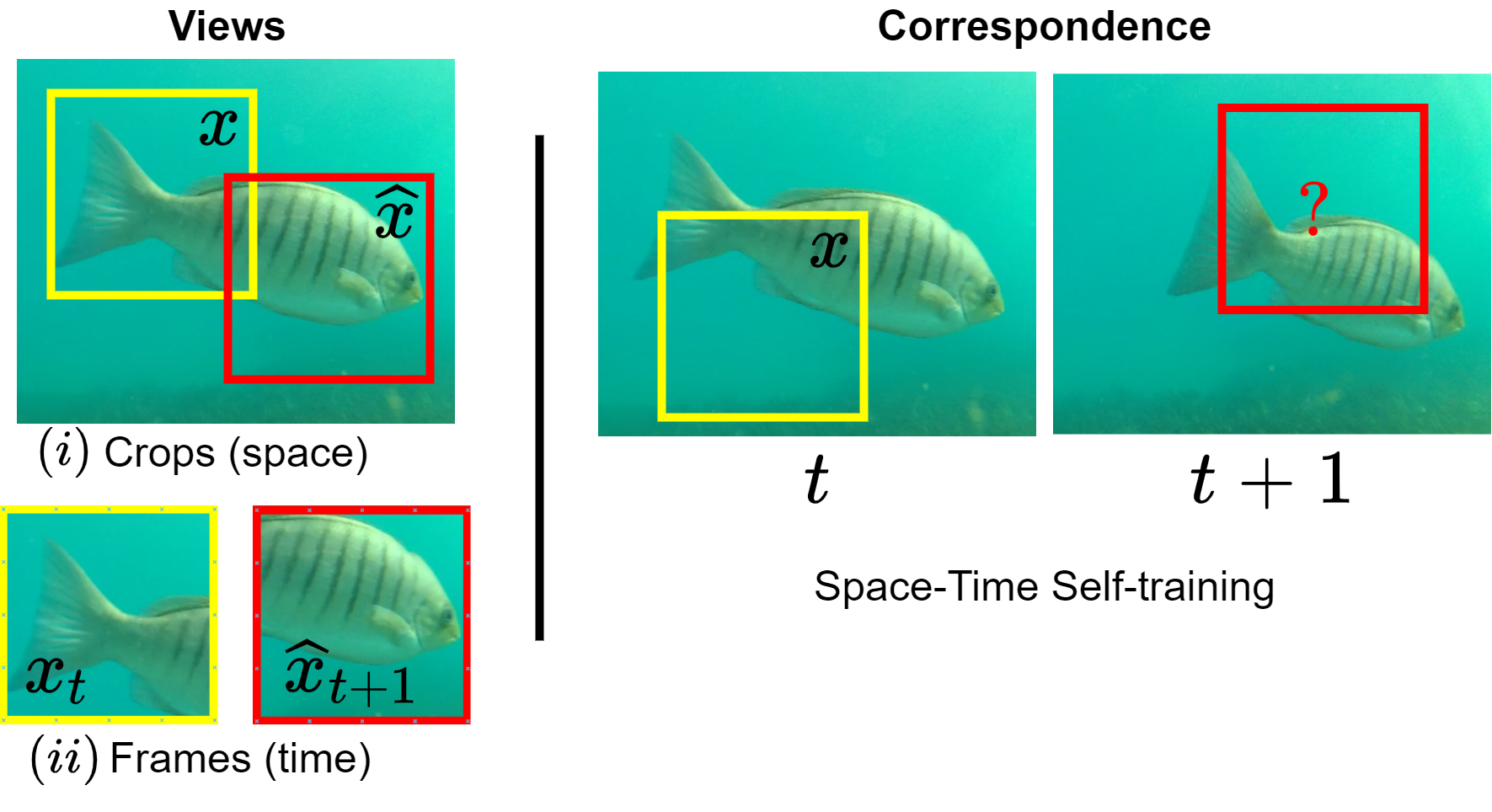}
\caption{Representation Learning as similarity across views by discriminating features \textit{(i)} spatially within individual frames and \textit{(ii)} temporally, to represent each frame in a video sequence in terms of the same feature set.}
\label{fig:8}
\end{figure} 
%%%%%%%%%%%%%%%%%%%%%%%%%%%%%%%%%%%%%%%%%%%%%%%%%%%%%%%%%%%%%%%%

\subsection{Loss Function} \label{secloss}
The training objective of this framework is to learn representations as similarity across views (Figure \ref{fig:8}). This is achieved by calculating pairwise affinities between features from the model's two branches and minimizing the distance of features extracted from temporally close frames to the anchors. The overall loss combines cross-view consistency and space-time self-training losses.

\paragraph{Cross-View Consistency}
The regularizing branch processes augmented versions of the original video frames, and the model should produce consistent segmentations regardless of random flipping or scaling. The cross-view consistency loss enforces this property:
\begin{equation}
\mathcal{L}_{CV} = - \sum_{i \in \mathcal{R}} \log \frac{\exp ( z_i \cdot \hat{z}_i / \tau ) }{\sum_{l \neq i} \exp (z_i \cdot \hat{z}_l / \tau) },
\label{eq:cv}
\end{equation}
where:
\begin{itemize}
    \item \(\mathcal{R}\) is the index set of features from the reference frames,
    \item \(\tau \in \mathbb{R}^{+}\) is a temperature hyperparameter,
    \item \(z\) represents features from the main branch,
    \item \(\hat{z}\) represents features from the regularizing branch.
\end{itemize}
This loss distinguishes the cosine similarity between corresponding features from non-corresponding pairs.

\paragraph{Space-Time Self-Training}
Pseudo labels are generated from the predominant anchor index for each feature in the regularizing branch. The space-time self-training loss minimizes the distance between features from the original view \(q\) (Equation \ref{eq:affinity}) and pseudo labels \(p\) (Equation \ref{eq:pseudo_label}):
\begin{equation}
\mathcal{L}_{ST} = - \sum_{i \not\in \mathcal{R}} \log \mathcal{T}(q_i, p_i),
\label{eq:st}
\end{equation}
where:
\begin{itemize}
    \item \(\mathcal{R}\) is the index set of features from the reference frames,
    \item \(\mathcal{T}(\cdot)\) is a random similarity transform aligning \(q\) and \(p\) after cropping and flipping.
\end{itemize}
This loss encourages increased cosine similarity of features to anchors and decreased similarity between anchors.

\paragraph{Final Loss}
The total loss combines the cross-view consistency and space-time self-training losses:
\begin{equation}
\mathcal{L} = \mathcal{L}_{CV} + \lambda \mathcal{L}_{ST},
\label{eq:loss_total}
\end{equation}
where \(\lambda\) is a hyperparameter weighting the contribution of \(\mathcal{L}_{ST}\).
%%%%%%%%%%%%%%%%%%%%%%%%%%%%%%%%%%%%%%%%%%%%%%%%%%%%%%%%%%%%%%%%
% \begin{figure}[t]
% \centering
% \includegraphics[width=0.98\linewidth]{fig_3.png}
% \caption{Sample image from each of the three utilised datasets. From left: DeepFish \cite{Saleh2020a}, Seagrass \cite{Ditria2021a} , and YouTube-VOS \cite{Xu2018b}}
% \label{fig:3}
% \end{figure}
%%%%%%%%%%%%%%%%%%%%%%%%%%%%%%%%%%%%%%%%%%%%%%%%%%%%%%%%%%%%%%%%

\subsection{Label Propagation} \label{seclab}
Label propagation is used to predict semantic labels for all video frames using only the ground truth from the first frame as input. This is essentially a classification task where each pixel in every frame of a video is assigned a label, given that only the first frame has ground truth labels available.
Following previous work, this study employs representation as a similarity function for k-Nearest Neighbour (KNN) prediction.
Algorithm \ref{code1} shows the label propagation procedure used in this work.

The mask $m_t$ for the current timestep $t$ is predicted using \emph{context} embeddings and masks from previous frames. The embedding for frame $t$ is obtained from the \texttt{CoaT Transformer} output.
Next, the cosine similarity of embedding $e_t$ with respect to all embeddings in context $\mathcal{E}$ is calculated. This approach is commonly used in correlation layers of optical flow networks.
Following this, local attention is computed in a single operation using \texttt{kNN-Softmax}.
Finally, the oldest entries in the mask $\mathcal{M}$ and embedding contexts $\mathcal{E}$ are updated by replacing them with $m_t$ and $e_t$.
This process is repeated for all remaining frames in the video clip.
Bilinear interpolation is used to resize the final object masks back to their original resolution.
%%%%%%%%%%%%%%%%%%%%%%%%%%%%%%%%%%%%%%%%%%%%%%%%%%%%%%%%%
\begin{algorithm}[ht] 
  \scriptsize
	% \small
    \caption{Label Propagation} \LinesNumbered \KwIn{Embeddings $\mathcal{E}$ and mask $\mathcal{M}$ from the first frame.} \KwOut{Mask $m_t$ prediction for timestep $t$.} \For{ $t$ and frame in (frames)}{ Computing embeddings $e_t$ at timestep $t$; Computing local spatial correlation between $\mathcal{E}$ and $e_t$; Computing softmax between K-Nearest Neighbors; Mask $m_t$ prediction for timestep $t$; updating $\mathcal{E}$ and $\mathcal{M}$; }
    \label{code1} 
    \end{algorithm}
%%%%%%%%%%%%%%%%%%%%%%%%%%%%%%%%%%%%%%%%%%%%%%%%%%%%%%%%%%

\section{Experiments} \label{secresult}
This section outlines the training and evaluation methodology for the self-supervised learning model for underwater video segmentation. Quantitative and qualitative results demonstrate the model's ability to generalize across diverse underwater habitats.

\subsection{Datasets} \label{secdata}
Experiments were conducted using three publicly available datasets: DeepFish \cite{Saleh2020a}, Seagrass \cite{Ditria2021a}, and YouTube-VOS \cite{Xu2018b}. 
% Figure \ref{fig:3} provides sample images from each dataset.

\begin{itemize}
    \item \textbf{DeepFish}: Contains 40k video frames captured in Full HD (1920 x 1080 pixels) across 20 habitats in tropical Australia.
    \item \textbf{Seagrass}: Includes 4280 video frames with 9429 annotations of \textit{Girella tricuspidata} in Australian estuaries, recorded using submerged action cameras.
    \item \textbf{YouTube-VOS}: Comprises 4453 YouTube video clips with pixel-level annotations for every 5th frame. Only 130 "fish" videos (4349 frames) were used for this study.
\end{itemize}
The feature extractor was trained on DeepFish and evaluated on Seagrass and YouTube-VOS.

\subsection{Data Augmentation} \label{secaug}
Training data was augmented using similarity transformations (random cropping and flipping), as illustrated in Figure \ref{fig:8}. These augmentations were applied to create variations for the regularizing branch, enabling pseudo-label generation. Experiments with additional augmentations (e.g., shearing, rotations, RGB-Shift, color jittering) showed no significant accuracy improvements and were computationally expensive. Thus, only random flips and cropping were used.

\subsection{Implementation Details}
\subsubsection{Model Training} \label{sectrain}
The Transformer-based feature encoder (Section \ref{seccoat}) served as the backbone network. For comparison, a ResNet-18 baseline was used, with strides removed from the \texttt{res3} and \texttt{res4} blocks. Both models were randomly initialized and trained on \(256 \times 256\) pixel inputs, achieved by scaling the shortest side to 256 and extracting random crops. Each forward pass used \(B \times T = 2 \times 5 = 10\) frames (2 video sets, 5 frames each).

Training used a learning rate of \(1 \times 10^{-3}\) and required approximately 300 epochs. Models were implemented in PyTorch and trained on an NVIDIA GeForce RTX 2080 Ti GPU (11 GB memory) using the Adam optimizer (\(\beta_1 = 0.5\), \(\beta_2 = 0.999\), \(\epsilon = 1.0 \times 10^{-8}\)). The same hyperparameters were applied to all models, though optimal configurations may vary by application. The training loop is outlined in Algorithm \ref{code2}.

\subsubsection{Inference} \label{secinfr}
During inference, dense correspondences for video propagation were computed using the learned encoder's representation (Section \ref{seccoat}, Algorithm \ref{code2}). Segmentation masks for entire video frames were predicted using label propagation (Section \ref{seclab}). Given the ground-truth mask for the first frame, labels were propagated to subsequent frames without additional annotations. Labels in the first frame were represented as one-hot vectors, while propagated labels were Softmax distributions.

%%%%%%%%%%%%%%%%%%%%%%%%%%%%%%%%%%%%%%%%%%%%%%%%%%%%%%%%%%
\begin{algorithm}[ht] 
  \scriptsize
	% \small
	\caption{Main Training Loop}
	\LinesNumbered
	\KwIn{Unlabeled video sequences.}
	\KwOut{Trained weights for the backbone network.}
	\For{each mini-batch}{
		Extract deep features of the video frames\;
		Regularising branch produces pseudo labels\;
		
		\For{each video in the mini-batch}{
			{\tt \textit {// Transformer-based encoder}}\\
			Extract feature embeddings (anchors) $k$\;
			Compute affinity to anchors ${q}$ (\cref{eq:affinity})\;
			Compute pseudo labels ${q}$ (\cref{eq:pseudo_label})\;

			{\tt \textit {// Loss Computation}}\\
			Compute Cross-view consistency $ \mathcal{L}_{\text{CV}} $ (\cref{eq:cv})\;
			Compute Space-time self-training $ \mathcal{L}_{\text{ST}} $ (\cref{eq:st})\;
			Compute total loss $ \mathcal{L} $ (\cref{eq:loss_total})\;
		}
		Back-propagate all the losses in this mini-batch\;
	} 
    \label{code2}
\end{algorithm} 

%%%%%%%%%%%%%%%%%%%%%%%%%%%%%%%%%%%%%%%%%%%%%%%%%%%%%%%%%%%%%%% 

\subsection{Evaluation Metrics}
Two commonly used metrics in video segmentation were employed to evaluate the model's performance:

\begin{itemize}
    \item \textbf{Jaccard's Index ($\mathcal{J}$)}: Measures the overlap between the predicted segmentation mask and the ground truth mask. It is calculated as:
$\mathcal{J}(A,B) = \frac{\mid A \cap B \mid} {\mid A \cup B \mid},$
    where \(A\) and \(B\) represent the predicted and ground truth masks, respectively.
    \item \textbf{Dice Coefficient ($\mathcal{F}$)}: Measures the overlap between the predicted and ground truth segmentations. It is calculated as:
$\mathcal{F} = \frac{2 \mid A \cap B \mid} {\mid A \mid + \mid B \mid},$
\end{itemize}
This study reports the mean average of \(\mathcal{J}\&\mathcal{F}\), as well as the mean and recall of \(\mathcal{J}_m\) and \(\mathcal{F}_m\), with an IoU threshold of 0.5.

\subsection{Compared Methods}
The proposed method was evaluated against five self-supervised video object segmentation methods and one fully-supervised method:
\begin{itemize}
    \item \textbf{Self-Supervised Methods}: CRW \cite{Jabri2020}, DenseFlow \cite{Araslanov2021}, MAST \cite{Lai2020a}, Colorize \cite{Vondrick2018tcv}, CorrFlow \cite{lai2019corrflow}.
    \item \textbf{Fully-Supervised Method}: FCN8 \cite{Shelhamer2017}, which uses true per-pixel class labels for training and handles imbalanced datasets effectively.
\end{itemize}
To ensure fairness, all self-supervised methods were trained on the DeepFish dataset using the authors' provided code and parameters.

%%%%%%%%%%%%%%%%%%%%%%%%%%%%%%%%%%%%%%%%%%%%%%%%%%%%%%%%%%%%%%%%
\begin{figure*}[htbp]
\centering
\includegraphics[width=0.48\textwidth]{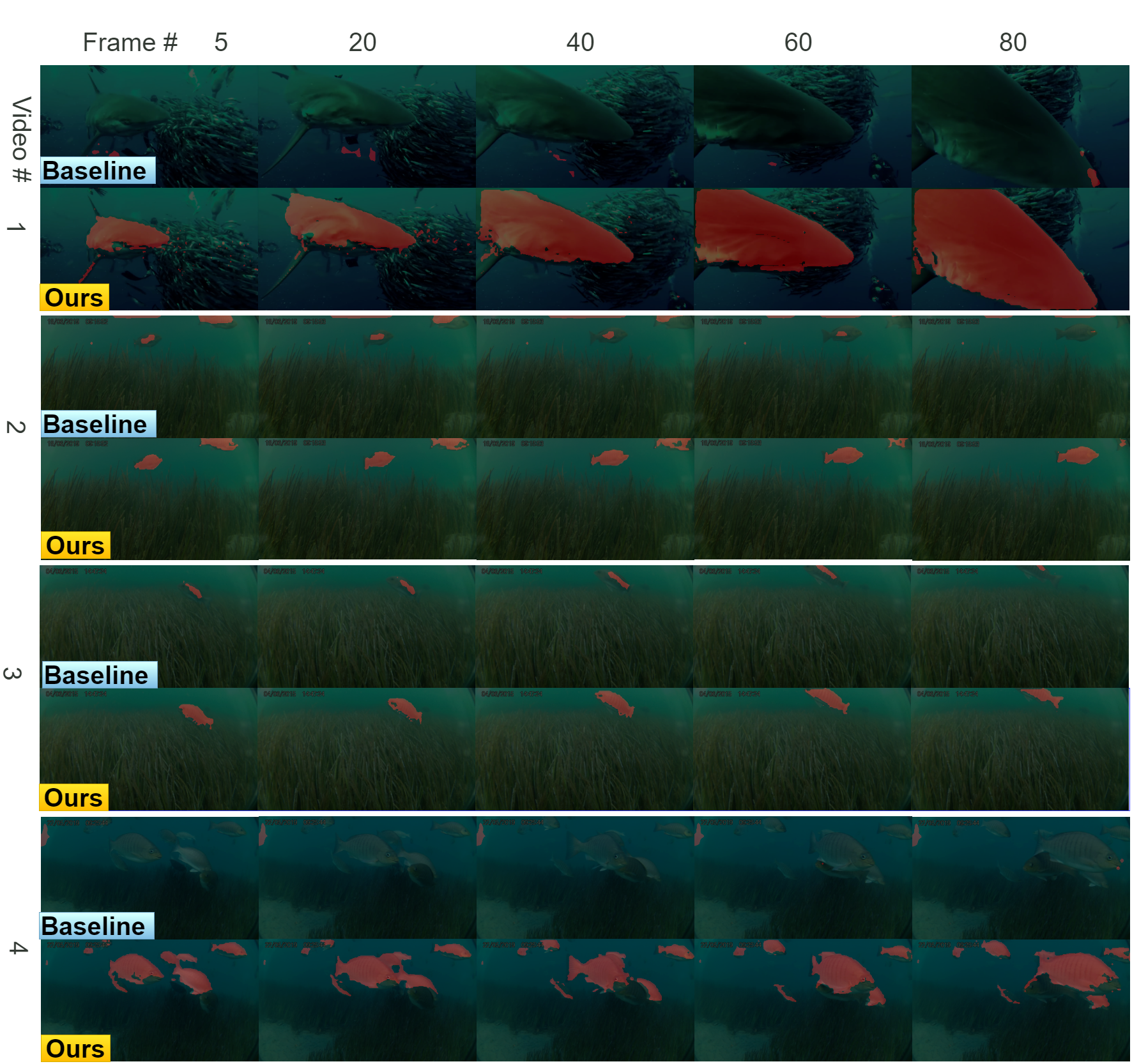}
\caption{\color{black}
Qualitative comparison between our model and a baseline \cite{Araslanov2021} model applied on the YouTube-VOS (rows 1 and 4) \cite{Xu2018b}, and Seagrass (rows 2 and 3) \cite{Ditria2021a}  datasets.
The representation learned by our model effectively distinguishes between objects and background ambiguity and is robust to occlusions.
}
\label{fig:5}
\end{figure*}
%%%%%%%%%%%%%%%%%%%%%%%%%%%%%%%%%%%%%%%%%%%%%%%%%%%%%%%%%%%%%%%% 

\subsection{Quantitative Comparisons} \label{secmetrc}
The results for the Seagrass and YouTube-VOS datasets are summarized in Tables \ref{tab:result1} and \ref{tab:result2}, respectively. The best result for each metric is highlighted in bold.

Compared to the second-highest performing method, our approach achieves higher \(\mathcal{J}\&\mathcal{F}_m\) scores by 4.5\% on Seagrass and 3.1\% on YouTube-VOS. This improvement is attributed to the self-attention mechanisms, which extract high-level spatial features and model dependencies across temporal steps, aiding in segmenting objects with large motions.

While the fully-supervised FCN8 method achieves higher \(\mathcal{J}\&\mathcal{F}_m\) scores, the proposed method does not require annotated frames, making it more practical for large-scale applications.

%%%%%%%%%%%%%%%%%%%%%%%%%%%%%%%%%%%%%%%%%%%%%%%%%%%%%%%%%%%%%%%%%%%
\begin{table*}[!htb]
\caption{\color{black}Performance Comparison on Seagrass \cite{Ditria2021a} dataset between our model and five state-of-the-art models 
(CRW  \cite{Jabri2020}
DenseFlow \cite{Araslanov2021}
MAST  \cite{Lai2020a}
Colorize  \cite{Vondrick2018tcv}
CorrFlow  \cite{lai2019corrflow}). }
\label{tab:result1}
\color{black}
\centering
\footnotesize\addtolength{\tabcolsep}{0pt}
\begin{spacing}{1.5}
\begin{tabular*}{\textwidth}{l @{\extracolsep{\fill}} cccccc}
\toprule
Method   & $\mathcal{J}$\&$\mathcal{F}$(Mean) $\uparrow$ & $\mathcal{J}$(Mean) $\uparrow$ & $\mathcal{J}$(Recall) $\uparrow$ & $\mathcal{F}$(Mean) $\uparrow$ & $\mathcal{F}$(Recall) $\uparrow$ \\ \midrule

CRW  \cite{Jabri2020}       &  43.2 &  38.9 & 40.4 &  46.2 & 50.8     \\
DenseFlow \cite{Araslanov2021}      & 45.5 & 40.2 & 41.0 & 50.7 & 54.7 \\
MAST  \cite{Lai2020a}       &  40.3 &  37.1 & 38.7 &  43.8 & 48.5     \\
Colorize  \cite{Vondrick2018tcv}       &  34.9 &  34.5 & 35.1 &  40.8 & 47.9     \\
CorrFlow  \cite{lai2019corrflow}       &  39.4 &  36.8 & 36.9 &  42.7 & 47.2     \\
Ours     & \textbf{50.0} & \textbf{41.5} & \textbf{43.3} & \textbf{58.1} & \textbf{65.4} \\
\cdashline{1-7}
Fully-supervised \cite{Shelhamer2017}     & 64.7 &   52.4 & 55.7 &  71.4 & 79.1 \\
\bottomrule
\end{tabular*}
\vspace{-8pt}
\end{spacing}
\end{table*}

%%%%%%%%%%%%%%%%%%%%%%%%%%%%%%%%%%%%%%%%%%%%%%%%%%%%%%%%%%%%%%%%%%%% 

%%%%%%%%%%%%%%%%%%%%%%%%%%%%%%%%%%%%%%%%%%%%%%%%%%%%%%%%%%%%%%%%%%%
\begin{table*}[!htb]
\caption{\color{black}Performance Comparison on YouTube-VOS \cite{Xu2018b} dataset between our model and five state-of-the-art models 
(CRW  \cite{Jabri2020}
DenseFlow \cite{Araslanov2021}
MAST  \cite{Lai2020a}
Colorize  \cite{Vondrick2018tcv}
CorrFlow  \cite{lai2019corrflow}). }
\label{tab:result2}
\centering
\color{black}
\footnotesize\addtolength{\tabcolsep}{0pt}
\begin{spacing}{1.5}
% \setlength{\tabcolsep}{.9pt}
% \resizebox{.90\linewidth}{!}{
\begin{tabular*}{\textwidth}{l @{\extracolsep{\fill}} cccccc}
\toprule
Method   & $\mathcal{J}$\&$\mathcal{F}$(Mean) $\uparrow$ & $\mathcal{J}$(Mean) $\uparrow$ & $\mathcal{J}$(Recall) $\uparrow$ & $\mathcal{F}$(Mean) $\uparrow$ & $\mathcal{F}$(Recall) $\uparrow$ \\ \midrule

CRW  \cite{Jabri2020}      &  59.9 &  58.6 & 71.5 &  57.8 & 68.7     \\
DenseFlow \cite{Araslanov2021}       & 60.2 & 60.9 & 72.7 & 59.5 & \textbf{70.0} \\
MAST  \cite{Lai2020a}     &  57.4 &  57.9 & 68.1 &  56.9 & 65.2     \\
Colorize  \cite{Vondrick2018tcv}       &  53.7 &  54.1 & 65.9 &  55.4 & 64.8     \\
CorrFlow  \cite{lai2019corrflow}       &  56.4 &  55.9 & 66.7 &  54.3 & 64.8     \\
Ours      & \textbf{63.3} & \textbf{63.9} & \textbf{74.0} & \textbf{62.7} & 69.6 \\
\cdashline{1-7}
Fully-supervised \cite{Shelhamer2017}     & 79.3 &   78.2 & 89.3 &  70.5 & 83.7 \\
\bottomrule
\end{tabular*}
% }
\vspace{-8pt}
\end{spacing}
\end{table*}

\subsection{Qualitative Results} \label{secqlt}
A qualitative comparison was conducted to visually assess the segmentation results and confirm the model's accuracy in identifying objects within images. The proposed model was compared against baseline models on the YouTube-VOS (rows 1 and 4) and Seagrass (rows 2 and 3) datasets, as depicted in Figure \ref{fig:5}. The results demonstrate that the proposed model effectively distinguishes between objects and backgrounds and outperforms the baseline in handling occlusions. Notably, the model excels at segmenting multiple overlapping fish in complex scenes, as shown in the bottom panel of Figure \ref{fig:5}.

The proposed model also exhibits stability and efficacy in accurately locating fish within long videos, even in complex scenes. For example, it successfully segmented up to frame number 80 based solely on the first frame. In contrast, the baseline method struggled with ambiguity between foreground objects and the background or complex transformations in the videos. The proposed method demonstrated a strong ability to differentiate pixels with similar intensities.

\begin{figure*}[htbp]
\centering
\begin{subfigure}[t]{0.48\textwidth}
    \centering
    \includegraphics[width=\textwidth]{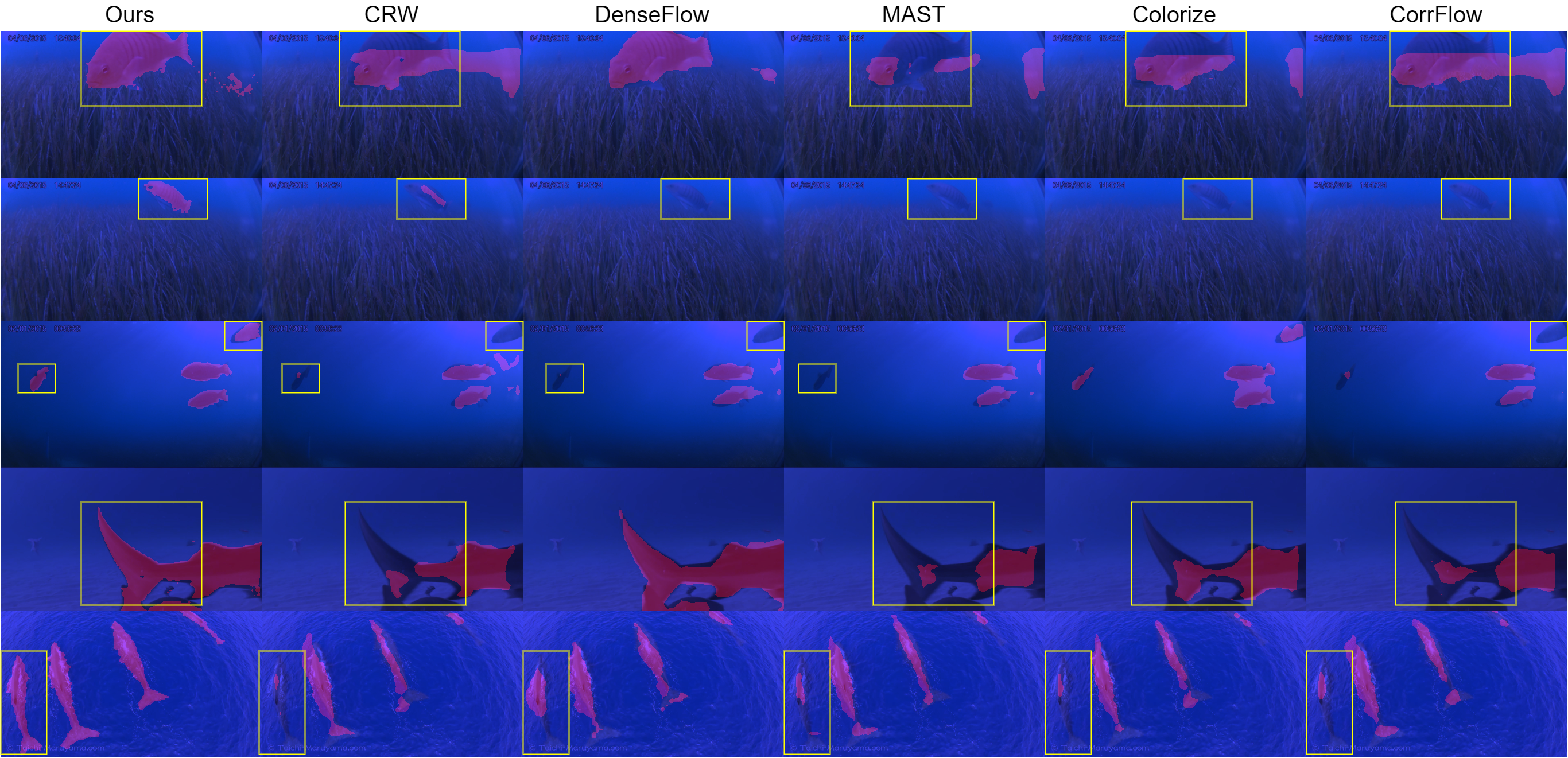}
    \caption{\color{black} 
    Qualitative comparison between our model and five state-of-the-art models 
    (CRW  \cite{Jabri2020}, DenseFlow \cite{Araslanov2021}, MAST  \cite{Lai2020a}, Colorize  \cite{Vondrick2018tcv}, CorrFlow  \cite{lai2019corrflow}) 
    applied on Seagrass \cite{Ditria2021a} (rows 1-3), and the YouTube-VOS \cite{Xu2018b} (rows 4 and 5) datasets. 
    The yellow rectangle highlights instances where the other methods did not correctly identify the fish body or a significant part of it.}
    \label{fig:2}
\end{subfigure}
\hfill
\begin{subfigure}[t]{0.48\textwidth}
    \centering
    \includegraphics[width=\textwidth]{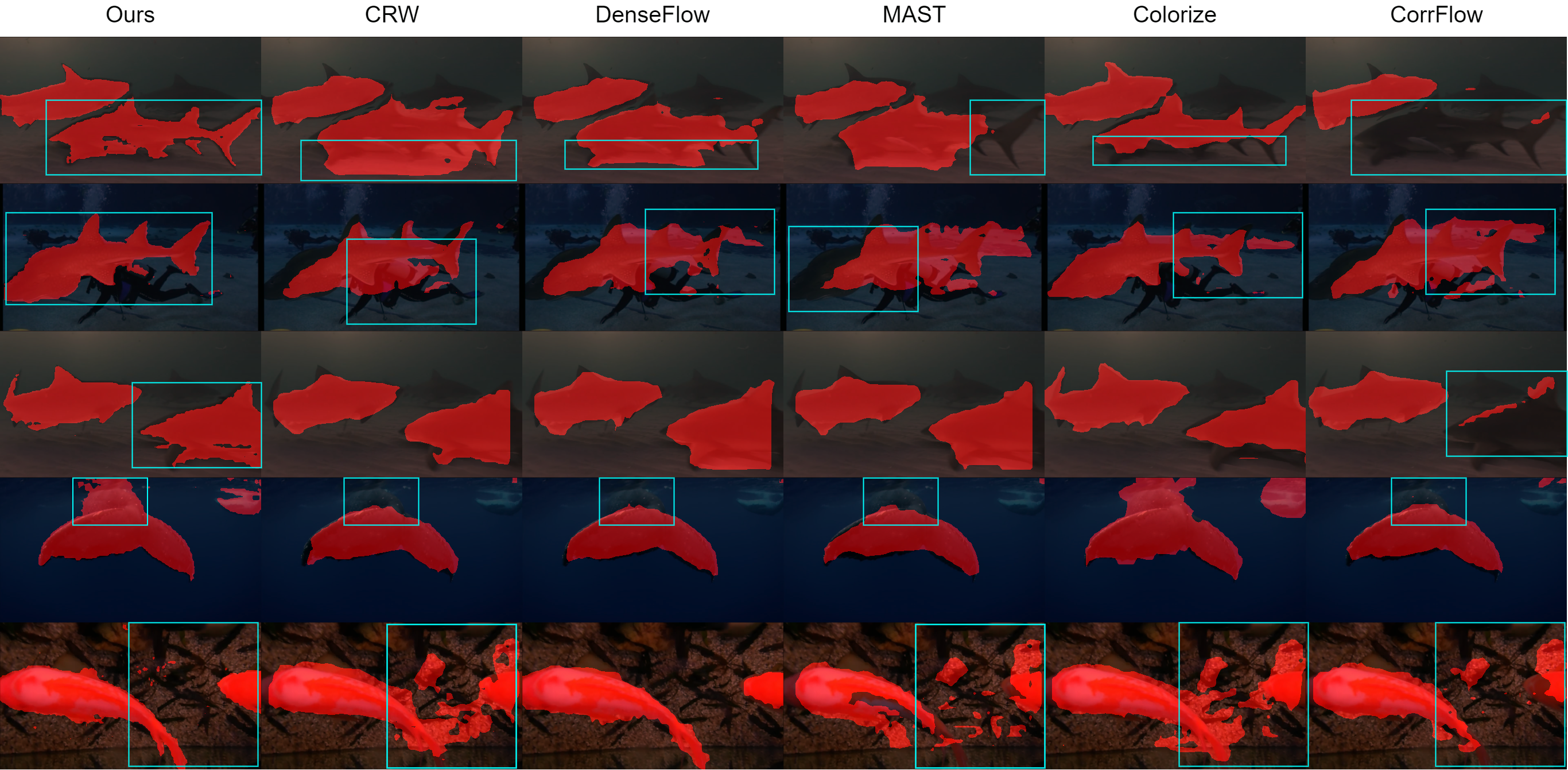}
    \caption{\color{black} 
    Qualitative comparison between our model and five state-of-the-art models 
    (CRW  \cite{Jabri2020}, DenseFlow \cite{Araslanov2021}, MAST  \cite{Lai2020a}, Colorize  \cite{Vondrick2018tcv}, CorrFlow  \cite{lai2019corrflow}) 
    applied on the YouTube-VOS \cite{Xu2018b} dataset. 
    The blue rectangle highlights instances where the other methods did not accurately identify the contour of the fish's body.}
    \label{fig:1}
\end{subfigure}
\caption{\color{black} 
Qualitative comparison between our model and five state-of-the-art models on the Seagrass and YouTube-VOS datasets. (a) Results from Seagrass and YouTube-VOS datasets. (b) Results from the YouTube-VOS dataset. Note: For better visibility of details, please view the figures online and zoom in.}
\label{fig:comparison}
\end{figure*}

Additionally, the model performed effectively on datasets containing very small objects, as seen in Videos 2 and 3 of Figure \ref{fig:5}. To further validate the proposed method, Figures \ref{fig:2} and \ref{fig:1} provide comparisons with five state-of-the-art self-supervised models (CRW, DenseFlow, MAST, Colorize, and CorrFlow) on the YouTube-VOS and Seagrass datasets, respectively.

%%%%%%%%%%%%%%%%%%%%%%%%%%%%%%%%%%%%%%%%%%%%%%%%%%%%%%%%%%%%%%%%
%%%%%%%%%%%%%%%%%%%%%%%%%%%%%%%%%%%%%%%%%%%%%%%%%%%%%%%%%%%%%%%%

\color{black}
%%%%%%%%%%%%%%%%%%%%%%%%%%%%%%%%%%%%%%%%%%%%%%%%%%%%%%%%%%%%%%%%
\subsection{Ablation Study} \label{secabl}

An ablation study was performed to thoroughly examine the impact of the proposed framework. This study involved comparing the baseline  models with our method.
To ensure a comprehensive analysis, various configurations of video features, and MLP layers were evaluated to determine the combination yielding the best performance. Only the results obtained from the most optimal configurations are reported.

Table \ref{tab:ablation} presents the segmentation accuracy of four different models based on the ${\mathcal{J}\&\mathcal{F}}_m$ metric, along with the baseline model and the proposed models.
In Table \ref{tab:ablation}, the second row corresponds to the Fast Fourier Convolution (FFC) model. The third row represents the Transformer for Semantic Segmentation, while the fourth row presents the MetaFormer-based architecture, PoolFormer. The fifth row represents the Cross-Covariance Image Transformer (XCiT).
Based on the evaluation summarised in Table \ref{tab:ablation}, the proposed  model significantly outperforms the baseline models and other meta-architecture methods across both datasets.

\begin{table}[ht]
\centering
\caption{Ablation study for other models on Seagrass \cite{Ditria2021a}  and YouTube-VOS \cite{Xu2018b} datasets.}
\label{tab:ablation}
% \begin{spacing}{.7}
% \setlength{\tabcolsep}{.1pt}
% \resizebox{\linewidth}{!}{
\resizebox{.90\linewidth}{!}{
\begin{tabular}{@{}lcccc@{}}
\toprule
  & \multicolumn{2}{c}{$\mathcal{J} {\&} \mathcal{F}_\textrm{mean}$} \\
  \cmidrule{2-4}
 Method   & Seagrass \cite{Ditria2021a} & YouTube-VOS \cite{Xu2018b}  
 \\ \midrule
Baseline  \cite{Araslanov2021}    & 45.5 & 60.2 \\
FFC \cite{chi2020ffc}             & 46.2 & 60.6 \\
Segmenter \cite{Strudel2021a}     & 41.5 & 52.7 \\
PoolFormer \cite{yu2021meta}      & 42.8 & 54.9 \\
XCiT \cite{Nouby2021xcit}         & 43.7 & 56.8 \\
% CNN                       & 59.5 & \textbf{70.0} \\
Ours        & \textbf{50.0} & \textbf{63.3} \\
% \cdashline{1-4}
% Fully-supervised \cite{Shelhamer2017}        & 64.7 & 79.3 \\
\bottomrule
\end{tabular}
}
% \end{spacing}
\end{table}

%%%%%%%%%%%%%%%%%%%%%%%%%%%%%%%%%%%%%%%%%%%%%%%%%%%%%%%%%%%%%%%%
\begin{figure*}[!ht]
\centering
\begin{subfigure}[t]{0.48\textwidth}
    \centering
    \includegraphics[width=\textwidth]{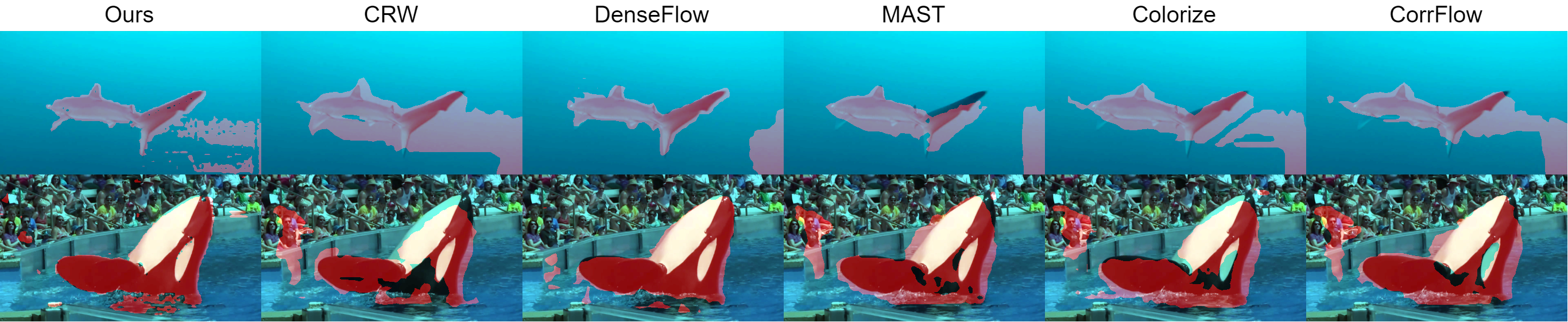}
    \caption{\color{black} Failure case of our model applied to one frame from the YouTube-VOS \cite{Xu2018b} dataset. Our model, similar to all other studied models, failed to differentiate between the fish and the background.}
    \label{fig:4}
\end{subfigure}
\hfill
\begin{subfigure}[t]{0.48\textwidth}
    \centering
    \includegraphics[width=\textwidth]{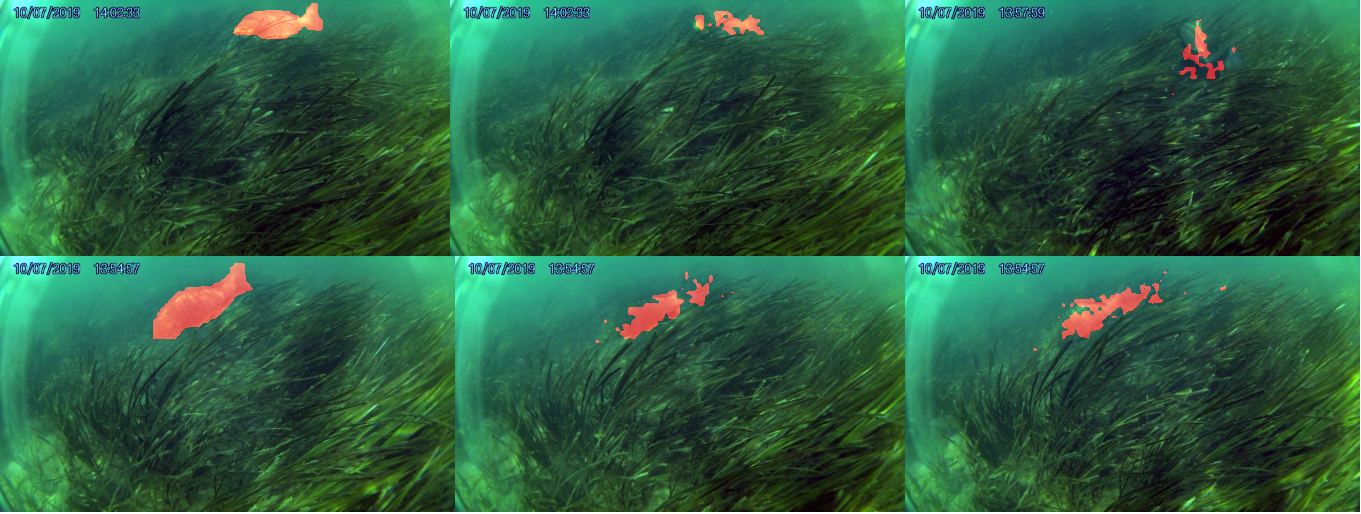}
    \caption{\color{black} Failure case of our model applied on a sequence of video frames from the Seagrass \cite{Ditria2021a} dataset. Our model failed due to heavy occlusion from seagrass.}
    \label{fig:10}
\end{subfigure}
\caption{\color{black} Failure cases of our model on different datasets. (a) YouTube-VOS dataset failure due to background similarity. (b) Seagrass dataset failure due to heavy occlusion. Note: For better visibility of details, please view the figures online and zoom in.}
\label{fig:failure_cases}
\end{figure*}
%%%%%%%%%%%%%%%%%%%%%%%%%%%%%%%%%%%%%%%%%%%%%%%%%%%%%%%%%%%%%%%%

%%%%%%%%%%%%%%%%%%%%%%%%%%%%%%%%%%%%%%%%%%%%%%%%%%%%%%%%%%%%%%%%
\subsection{Failure Case} \label{secfail}
During the experiments, it was observed that in a few instances, portions of the background were mistakenly segmented as fish. Two such examples are illustrated in Figure \ref{fig:4}. It is suggested that incorporating additional modalities, such as depth or motion information, could potentially enhance the attention model's performance in scenarios where the input contains objects visually similar to the background. This could be particularly beneficial when objects and backgrounds share similar colour or texture characteristics.
Another observation was that the model struggled to accurately segment the entire fish body in cases of heavy occlusion from seagrass. Figure \ref{fig:10} showcases instances of this issue.

% \color{blue}
\section{Limitations} \label{seclimitations}
While this study introduces a novel self-supervised learning approach for underwater fish segmentation, several limitations must be acknowledged. First, the model was trained on the \textit{DeepFish} dataset, which is specific to tropical Australian coastal environments. Although it generalized well to the Seagrass and YouTube-VOS datasets, its performance in other habitats or on rare fish species remains unvalidated. Second, the model occasionally missegmented background elements resembling fish in color or texture and struggled with heavy occlusions (e.g., seagrass). Performance in extreme conditions like poor lighting, turbidity, or variable water clarity was not extensively tested, which could limit its applicability in such scenarios. Third, errors in the initial ground-truth segmentation could propagate to subsequent frames due to the label propagation approach. Finally, while computationally efficient, the suitability of the model for low-power or mobile edge devices was not evaluated, and further optimisation may be required for deployment in resource-constrained environments.
% \color{black}
% \section{Limitations of the Work} \label{seclimitations}

%%%%%%%%%%%%%%%%%%%%%%%%%%%%%%%%%%%%%%%%%%%%%%%%%%%%%%%%%%%%%%%%
% \color{blue}
\section{Conclusion} \label{secconc}
This study introduced a self-supervised learning approach for segmenting fish in underwater videos, eliminating the need for manual annotation and enabling efficient data preparation. The model achieved promising results, with segmentation accuracy comparable to fully supervised methods in specific scenarios, though supervised methods may outperform it under ideal conditions. While the approach generalized well to the Seagrass and YouTube-VOS datasets after training on DeepFish, further validation in more diverse underwater environments is needed. The model's performance is currently limited in cases of severe occlusion, such as dense seagrass, and its suitability for low-power devices remains unexplored. Overall, this method provides a viable alternative to fully supervised approaches, particularly where manual annotation is impractical. Future work could focus on improving robustness to occlusions and extending the model to tasks like fish species identification and segmentation of other underwater objects.
% \color{black}

%%%%%%%%%%%%%%%%%%%%%%%%
\section*{Acknowledgement}
% \paragraph{Acknowledgments}
This research is supported by an Australian Research Training Program (RTP) Scholarship and Food Agility HDR Top-Up Scholarship. D. Jerry and M. Rahimi Azghadi acknowledge the Australian Research Council through their Industrial Transformation Research Hub program.

\section*{Data availability statement}
The datasets generated during and analysed during the current study are  publicly available.

% \section*{Conflict of interest statement}
% The authors declare that they have no known competing financial interests or personal relationships that could have appeared to influence the work reported in this paper.

% \section*{Funding}
% No funding was received to assist with the preparation of this manuscript.
% The authors have no relevant financial or non-financial interests to disclose.

% \section*{Ethics approval statement}
% This article does not contain any studies with human participants or animals performed by any of the authors. Informed consent was obtained from all individual participants included in the study.

% \section*{Funding}
% This project was supported by CRC project funding from the Department of Industry, Innovation and Science, Commonwealth of Australia. Mainstream Aquaculture Group partnered with James Cook University, The University of Melbourne and four commercial farming operations to deliver this project. 

% \bibliographystyle{IEEEtran}
% \bibliographystyle{cas-model2-names}
% \bibliographystyle{IEEEtranN}
% \bibliographystyle{IEEEtranSN}
% \bibliographystyle{apalike}
% \bibliography{sn-bibliography}% common bib file
\bibliography{references}% common bib file
% \bibliography{referenes}
% \bibliography{references}% common bib file


\end{document}